\documentclass[journal,a4paper]{IEEEtran}
\usepackage[utf8]{inputenc}
\usepackage{graphicx}
\usepackage{stfloats}
\usepackage{amsmath}

\usepackage{color}
\usepackage{amsfonts}
\usepackage{todonotes}
\usepackage{algorithm,algorithmic}
\usepackage{multirow,url,subcaption}
\usepackage{array}
\usepackage{etoolbox}
\usepackage{ifthen}

\makeatletter
\patchcmd{\@makecaption}
  {\scshape}
  {}
  {}
  {}
\makeatother

\DeclareMathOperator*{\argmax}{argmax}
\usepackage{amssymb}

\begin{document}
%
\title{Patch based adaptive temporal filter and residual evaluation}
%
%
%

\author{Weiying Zhao, Paul Riot, Charles-Alban Deledalle,  Henri Ma{\^\i}tre, Jean-Marie Nicolas, \\Florence Tupin,~\IEEEmembership{Senior Member,~IEEE}


\thanks{W. Zhao, H. Ma{\^\i}tre, J.-M. Nicolas, F. Tupin are with LTCI, T\'el\'ecom ParisTech, Universit\'e Paris-Saclay, 75013 Paris, France (e-mail: name.surname@telecom-paristech.fr)}
\thanks{C.-A. Deledalle is with IMB, CNRS, Univ. Bordeaux, Bordeaux INP,
F-33405 Talence, France (e-mail: charles-alban.deledalle@math.u-bordeaux.fr)}
\thanks{L. Denis is with Univ Lyon, UJM-Saint-Etienne, CNRS, Institut d~Optique Graduate School, Laboratoire Hubert  Curien UMR 5516, F-42023, SAINT-ETIENNE,
France (e-mail: loic.denis@univ-st-etienne.fr)}
}

\maketitle

\begin{abstract} 
In coherent imaging systems, speckle is a signal-dependent noise that visually strongly degrades images' appearance. A huge amount of SAR data has been acquired from different sensors with different wavelengths, resolutions, incidences and polarizations. 
We extend the nonlocal filtering strategy to the temporal domain and propose a patch-based adaptive temporal filter (PATF) to take advantage of well-registered multi-temporal SAR images.  A patch-based generalised likelihood ratio test is processed to suppress the changed object effects on the multitemporal denoising results. 
Then, the similarities are transformed into corresponding weights with an exponential function. The denoised value is calculated with a temporal weighted average.
Spatial adaptive denoising methods can improve the patch-based weighted
temporal average image when the time series is limited. The spatial adaptive denoising step is optional when the time series is large enough.
Without reference image, we propose using a patch-based auto-covariance residual
evaluation method to examine the ratio image between the noisy and denoised images and look for possible remaining structural
contents. It can process automatically and does not rely on a supervised selection of
homogeneous regions. It also provides a global score for the whole image.  Numerous results demonstrate the effectiveness of the proposed time series denoising method and the usefulness of the residual evaluation method. 

\end{abstract}


%
\IEEEpeerreviewmaketitle

\section{Introduction}
\IEEEPARstart{S}{ynthetic} aperture radar (SAR) imaging is a widely used remote measuring method. However, the inherent speckle attached to any coherent imaging system affects the analysis and interpretation of SAR images. Therefore, image filtering plays a significant role in successfully exploiting SAR images.
The main problems of SAR image denoising are spatial resolution preservation, edge preservation, texture areas denoising, bias cancelling and strongly reflecting target preservation. 

Spatial denoising methods only pay attention to one image using spatial information, such as weighted pixels to estimate the noise-free pixel values.  This can induce biased denoised values when participating pixel candidates are poorly selected. In addition, even using powerful spatial denoising approaches (such as SAR-BM3D \cite{parrilli2012nonlocal} and NL-SAR \cite{DDF+15}), the tiny structures can be damaged.
Noise is randomly distributed in SAR images. 
Most speckle reduction methods are based on the statistical characteristics of the noise, such as multiplicative gamma noise, Rayleigh-Nakagami noise, additive Gaussian noise, Poisson noise,  etc.   The popular used spatial denoising methods can be divided into four families: Bayesian methods in the spatial domain, Bayesian methods in  the transform domain,  selection-based filtering and other 
non-Bayesian approaches.

 With well-registered multi-temporal images, both spatial and temporal
information can be exploited in the denoising process, which gives the potential to provide better denoising results than using only a single image. Most multitemporal despeckling methods are extended from the single-image denoising techniques. 

To overcome the resolution broadening of the spatial denoising approach, Lee et al.  \cite{LGaM91}  take the multi-channel and multi-frequency information into account during the filtering. An adaptive moving rectangle window is utilized to estimate the correlation coefficient. 
Quegan and Yu \cite{quegan2001filtering} propose different ways of dealing with correlated and uncorrelated multitemporal images, along with recursive implementations. This method has been successfully applied to forest mapping \cite{quegan2000multitemporal}.
The significant advantage of these filters is that temporal SAR images acquired using different sensors can be jointly filtered.
However, a box window may destroy fine features, cause spatial blurring, and introduce wrong results near the bright points. 
To suppress the negative effects of changed points,  \cite{LAT+14, le2015change, SDT+14} propose to use only the unchanged candidates to complete the filtering. This approach can effectively suppress the changed bright points effect. However, the denoising results are highly influenced by the change detection threshold. An improper threshold may lead to bias denoising results, especially in the seasonal changed areas whose changes are hard to detect.

 In \cite{ciuc2001adaptive}, only the adjacent similar pixels with the same distribution as the target pixel are used to estimate the noise-free value. It extends the spatially based denoising method to the spatial-temporal domain. This method can preserve fine textures.
 
 SAR images usually exist with self-similarities. This makes it possible to use non-local filtering. Non-local means have been applied to multitemporal SAR image denoising, such as 2SPPB  \cite{SDT+14}, Nonlocal temporal filter (NLTF) \cite{chierchia2017multitemporal}, Multitemporal SAR-BM3D 
\cite{chierchia2017multitemporal} and RABASAR \cite{zhao2019ratio}. All these methods used the generalized likelihood ratio (GLR) \cite{deledalle2009iterative} to calculate the patch similarities.
These approaches can provide better denoising results and be recognized as state-of-the-art despeckling methods.

In addition, temporal denoising can also be processed in a transformed domain.  After performing logarithm and  Discrete Cosine Transformation, Coltuc et al. \cite{CTB+00} proposed applying refined Lee filtering to denoise the image in the spatial domain. Furthermore, MSAR-BM3D \cite{chierchia2017multitemporal} pursued denoising in the wavelet transformation domain.

These methods exploit different search approaches to select similar points, such as fixed rectangle window size, adaptive 3D window size, 3D selected patches, etc.
However, large time series will increase computational complexity, leading to higher computing storage requirements.

Multitemporal denoising methods could take advantage of more and more available SAR images to solve the spatial denoising problems for the benefit of resolution preservation. We aim to improve filtering results by exploiting temporal information. In addition, when enough images are available,  the temporal average image carries significant information.

In this paper, we extend the non-local method to the temporal domain. After acquiring different patch similarities in the time series, we transfer them to corresponding weights. Then, the denoised value is obtained through the temporal weighted average. We propose a residual evaluation method to evaluate the denoising performance, which does not have ground truth or a noise-free image. 
 
 \section{Patch-based adaptive temporal filter}
 \subsection{Similarity test}

Resolution cells comprise many scatterers with different phases, leading to interferences and a noise-like effect known as speckle. To limit the effect of speckles, usually $L$ pixels are averaged ($L$ is called the number of looks).
Under Goodman's hypothesis \cite{goodman2007speckle}, the fully developed intensity speckle follows a gamma distribution $\mathcal{G}[u,L]$ depending on the number of looks $L$ and the mean reflectivity $u$ of the scene:
\begin{equation}
    \mathcal{G}[u,L](y)=\frac{L}{u\Gamma(L)}\bigg(\frac{Ly}{u}\bigg)^{L-1} e^{-\frac{Ly}{u}}
    \label{eq:Gamma}
\end{equation}

 Speckle in coherently processed SAR data acts like a multiplicative noise, and the speckle model can be expressed as \cite{lee1981speckle}:

\begin{equation}
    \emph{y}=uv 
    \label{eq:y}
\end{equation}
 \begin{equation}
   \mathbb{E}[y] =u
\end{equation}
\begin{equation}\label{eq:variance}
     \mathrm{Var}[y]=u^2/L
\end{equation}
where $v$ follows a gamma distribution $\mathcal{G}[1,L]$, $\mathrm{Var}[\cdot]$ represents the variance operator. With the increase of the number of looks $L$, the variance ${{u^2}/{L}}$ decreases. A multiplicative signal model can describe both intensity and amplitude data. 

The similarity of two intensity observations can be measured as to whether there are changes.
Due to the multiplicative noise influence of coherent SAR images, hypothesis tests (such as likelihood ratio tests) are often used \cite{radke2005image}. 
In statistics, change detection can be considered as the comparison of two hypotheses $H_0$ and $H_1$, which corresponds to the null and alternative hypothesis, respectively.

$H_0$: $u_1=u_2=u_{12}$ (no change)

$H_1$: $u_1 \ne u_2$ (change)

The likelihood ratio test (LRT) is based on the likelihood ratio of observations $y_1$, $y_2$, which is defined as:

\begin{equation}
    LRT(y_1,y_2)=\frac{P(y_1,y_2 \mid u_{12},H_0 )}{P(y_1 \mid u_{1},H_1) P(y_2 \mid u_{2},H_1)}
\end{equation}
where ${u}_1$ and $u_{2}$ are the noise-free intensity values corresponding to $y_1$ and $y_2$ with associated ENL ${L}_1$ and ${L}_2$, $u_{12}$ is the noise-free value when there is no change between the intensity values.
Considering that $L$ look intensity data follows a gamma distribution, we have:

\begin{equation}
\begin{aligned}
P(y_1, y_2 \mid u_{1},u_{2},H_1)=
   & \frac{1}{\Gamma({L_1})}\frac{1}{\Gamma({L_2})}
    \bigg(\frac{L_1}{u_1}\bigg)^{L_1}
    \bigg(\frac{L_2}{u_2}\bigg)^{L_2}\\
   & y^{L_1-1}_1y^{L_2-1}_2
    \exp\bigg(-\frac{y_1L_1}{u_1}-\frac{y_2L_2}{u_2}\bigg)
\end{aligned}
\end{equation}
and under the hypothesis $H_0$, with $u_1=u_2=u_{12}$ we get:

\begin{equation}
\begin{aligned}
    P(y_1,y_2 \mid u_{12},H_0)=
    &\frac{1}{\Gamma({L_1})}\frac{1}{\Gamma({L_2})}
    \frac{L_1^{L_1}L_2^{L_2}}{u_{12}^{L_1+L_2}}
    y^{L_1-1}_1y^{L_2-1}_2\\
    &\exp\bigg(-\frac{y_1L_1+y_2L_2}{u_{12}}\bigg)
\end{aligned}
\end{equation}

These joint probability density functions allow us to use the hypothesis testing framework to describe their similarities.

Since ${u}_1$,  $u_{2}$ and $u_{12}$ are not available, they can be replaced by their maximum likelihood estimations under $H_0$:  
\begin{equation}\label{eq:MLestimationU1}
    \hat{u}^{ML}_1=y_1
\end{equation}
\begin{equation}
    \hat{u}^{ML}_2=y_2
\end{equation}
\begin{equation}\label{eq:MLestimationU12}
    \hat{u}^{ML}_{12}=\frac{{L}_1 y_1+{L}_2y_2}{{L}_1+{L}_2}
\end{equation}

Then, the generalized likelihood ratio test ($\text{GLRT}$) is given by: 

\begin{equation}\label{eq:GLRT}
\begin{aligned}
        \text{GLRT}(y_1,y_{2})
        &=({L}_1+{L}_2)^{{L}_1+{L}_2} \frac{{ y_1 }^{{L}_1}{y_2 }^{{L}_2}}{{( {L}_1y_1+{L}_2y_2 )}^{{L}_1+{L}_2}}
\end{aligned}
\end{equation}

Based on the maximum likelihood estimation results (Eq.(\ref{eq:MLestimationU1}-\ref{eq:MLestimationU12})), the logarithmic version of $\text{GLRT}$ is given by \cite{SDT+14}:
\begin{equation}\label{eq:logGLRT}
\begin{aligned}
    S_{GLR}(y_1,y_{2})
    &=L_1\log \frac{L_1 y_1+L_2y_2}{y_1(L_1+L_2)} + L_2\log \frac{L_1 y_1+L_2y_2}{y_2(L_1+L_2)}
\end{aligned}
\end{equation}

Suppose the corresponding pixels have the same ENL $L_1=L_2={L}$,  the simplified logarithmic version of $\text{GLRT}$ method \cite{deledalle2009iterative,SDT+14} turns out to be: 
\begin{equation} \label{eq:glr0}
    S_{GLR}(y_1,y_{2})=2{L}\log{\bigg(\sqrt{\frac{y_1}{y_2}}+
    \sqrt{\frac{y_2}{y_1}}}\bigg) -2{L}\log{2}
\end{equation}

 \subsection{Temporal weighted average}
 Patches are always used to suppress the noise effect.
 Note that similar patches could have structured information (edge, texture, $\cdot\cdot\cdot$), which is not considered during the temporal averaging.
Block similarity estimated based on generalized likelihood ratio test \cite{deledalle2009iterative, parrilli2012nonlocal} is equal to:
\begin{equation}\label{eq:Glrdistance}
    d_1[\emph{y}_t,\emph{y}_{t'}]=(2L-1)
    \sum_{k}\log
    \bigg(
    \sqrt{\frac{\emph{y}_t(s+k)}{\emph{y}_{t'}(s+k)}}+
    \sqrt{\frac{\emph{y}_{t'}(s+k)}{\emph{y}_t(s+k)}}
    \bigg) 
\end{equation}
where $s$ is the centre locations of compared patches.

The patch-based weighted average is exploited to suppress the effect of changed points.
In a noniterative version, the weights are calculated using:
 \begin{equation}
    \omega(y_t,y_{t'}) =
    \left\{
        \begin{array}{cc}
                0, & d_1[\emph{y}_t,\emph{y}_{t'}] \ge \tau_2\\
                \exp(-d_1[\emph{y}_t,\emph{y}_{t'}]/h)&\tau_1 < d_1[\emph{y}_t,\emph{y}_{t'}] < \tau_2\\
               \omega_{max}, &  d_1[\emph{y}_t,\emph{y}_{t'}] \le \tau_1 \\
        \end{array} 
    \right.
\end{equation}
\begin{equation}
    \omega_{max}=\argmax\limits_{t'} \exp(-d_1[\emph{y}_t,\emph{y}_{t'}]/h), \quad   \tau_1 < d_1[\emph{y}_t,\emph{y}_{t'}] < \tau_2
\end{equation}

\begin{equation}
    h \triangleq q- \mathbb{E}[c(v_{\Delta_{s}}, v_{\Delta_{s'}})]
\end{equation}

\begin{equation}
    q=\mathbb{F}^{-1}_{c(v_{\Delta_{s}}, v_{\Delta_{s'}})}(\alpha)
\end{equation}
where $h$ is a smoothing parameter defined as introduced in \cite{deledalle2009iterative}, $d_1$ is given by equation (\ref{eq:Glrdistance}), $q$ represents the $\alpha$-quantile and $\mathbb{F}$ is the cumulative distribution function. The larger $h$ will bring a smoother temporal average. $\tau_1$ and $\tau_2$ are two thresholds that separate different change magnitude patches.

  In practice, the original images or the dissimilarities can be multilooked to suppress the speckle corruption.
Large dissimilarities express that changes may have happened between the two patches. Some exponential kernels could be used to adjust the large distances to nearly zero weights \cite{kervrann2006optimal}. 
Under the no-change hypothesis, we propose to use the Monte Carlo method to define the unchanged (noise-free values are the same) and changed (changed to other objects) threshold, with probability quantiles equal to $8\%$ and $92\%$,  respectively.

\begin{figure*}[t]
\graphicspath{{results/}}
   \centering
\begin{tabular}{c}
\includegraphics[width=14cm]{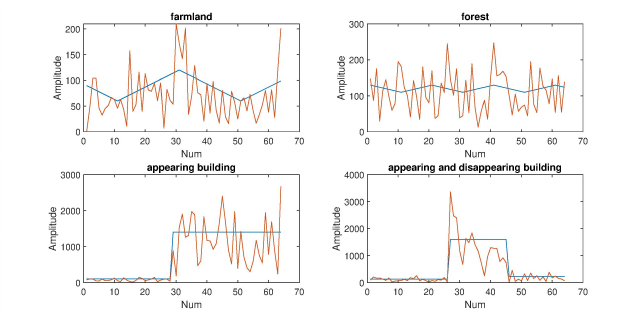}\\
(a) Noise-free (blue) and noisy (brownish yellow) time series.\\
\includegraphics[width=14cm]{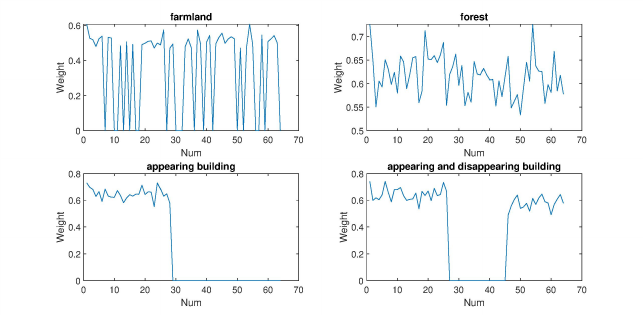}\\
(b) Different time series weights concerning (a).\\
\end{tabular}
\caption{Different time series and patch-based weight comparison. The self-similarity value of the reference point is set equal to the maximum similarity value with others.  $h$ is set equal to 2 during the transformation.}
	\label{fig:WeightTemporal}
\end{figure*}

With $M$ temporal images, the temporal weighted average is performed as follows:

\begin{equation}\label{eq:PATFweight}
    \hat{u}_t(s)=\frac{1}{\sum_{t'=1}^{M}\omega(y_t,y_{t'})}\sum_{t'=1}^{M}\omega(y_t,y_{t'})\emph{y}_{t'}(s) 
\end{equation}

Four kinds of object changes are taken as examples to show the dissimilarity to weight transformations (Fig.\ref{fig:WeightTemporal}). 
We only use patch-weighted temporal averaging for the long time series, such as more than 300 images, to obtain the denoised values. When only a few images are available, we can use spatial adaptive denoising to acquire better denoising results.   SAR-BM3D is chosen as an example of the despeckling scheme.  We use the PATF to represent the method when applying SAR-BM3D on the temporal weighted average results.

 \section{Residual evaluation}
Qualitative and quantitative measures of speckle reduction performances are challenging, especially when noise-free data are unavailable. Visually checking the despeckling
results is an immediate and subjective way for quality evaluation, but it is limited by human vision ability. To overcome this limitation, we propose a residual evaluation method.

The evaluation method proposed here follows the idea presented in \cite{riot2017correlation}, which 
examines the residual image and looks for possible remaining structural elements in this residual image. Unlike maximum ENL estimation or $\alpha\beta$ estimation \cite{gomez2016new} method, this method is automatic and does not rely on a supervised selection of homogeneous regions. It also provides a global score for the whole image. 


The ratio between noisy data $y$ and denoised data $\hat{u}$  is computed through:
\begin{equation}
    {R}=\frac{y}{\hat{u}}
\end{equation}

After denoising, the ratio $R$ is expected to contain pure speckle, except in regions where some structures were lost during the denoising procedure. In \cite{riot2017correlation}, the remaining structures are evaluated with autocorrelation. In our case, autocovariance is used for the evaluation of the residual. 
The autocovariance estimator $C_{P_{s+k}}$ on a noisy patch $P$ of size $W_{s}$, centered at location $s$, for all surrounding patches $P(s+k)$ is equal to:
\begin{equation}\label{autocorrelation}
    C_{P_{s+k}}=P(s)P(s+k)
\end{equation}

The mean value of the ratio is removed during the calculation.
We propose to estimate a normalized residuals autocovariance $ C^{norm}_{P_s}$  on each patch
$P_s$ of the residual as:
\begin{equation}
   C^{norm}_{P_s}=\frac{N^2_1}{N^2_1-1}\cdot
    \frac{
    \|C_{P_s} \|^2_2-\|C_{P_s}(0,0) \|^2_2}
    {\sum^{}_{s+k\in W_s}{(R^2_{s+k}-\mathbb{E}[{R}])} }
\end{equation}
with $N^2_1$ the patch size and $\mathbb{E}[{R}]=1$,  because of the unit mean of Gamma distribution noise. 
Then, we can aggregate the $C^{norm}_{P_s}$ covariances computed
over each patch to obtain a quality map at each pixel $s'$:

\begin{equation}
    W_{map}(s')=\frac{\sum_{}{w}(s,s')C^{norm}_{P_{s,s'}}
    }{\sum_{}{w}(s,s')}
\end{equation}
where ${w}(s,s')=1$ if the pixel in position $s'$ of the residual is used in the patch $P_s$, $C^{norm}_{P_{s,s'}}$ represents the autocovariance value of $s'$ in the patch $P_s$.
 Then, a global score can be obtained with an average:
\begin{equation}\label{scorenormalized}
    W_{score}=\frac{1}{N}\sum_{(s')}W_{map}(s')
\end{equation}
where $N$ is the number of pixels in the image.

In practice, we will limit the autocovariance to displacements of one pixel in both vertical and horizontal directions to speed up the process, following the recommendation of \cite{dabov2007image}.
In the quality map $W_{map}(s')$, low values inform where an efficient denoising has been obtained.
The residuals evaluation method does not need any ground truth. Therefore, it is suited for evaluating the denoising of real SAR images.

\section{Experimental results and discussion}

We will use simulated and SAR data to compare the proposed method and the influence of the nonlocal means estimation. The effectiveness of patch-based weighted temporal averaging is proved using long time-series Sentinel-1 GRD images. In addition, the proposed methods will be compared with some state-of-the-art multitemporal denoising approaches (such as 2SPPB \cite{SDT+14}, MSAR-BM3D \cite{chierchia2017multitemporal} and RABASAR \cite{zhao2019ratio}).

Because fine textures are present in SAR images, using the spatial average is
often unsuitable. Large search windows smear fine edges, point targets and fine features, and the spatial similarity points acquired by NLTF may belong to different
objects. An adaptive spatial averaging
method, such as SAR-BM3D, can be used to get a better reflectance estimation. Here, we call the improved method adaptive nonlocal temporal filter (ANLTF). 
ANLTF used the weighted average value of the search window. PATF used the weighted average value of the time series.
The main difference between PATF and UTA, NLTF and ANLTF is the noise-free value estimation method. The four compared methods are shown in Tab.\ref{tab:noisefreeEst}. UTA \cite{LGaM91,quegan2001filtering} method used the average value of the rectangular window. NLTF \cite{chierchia2017multitemporal} only averages similar points estimated using generalized likelihood ratio test \cite{deledalle2012compare}.

\begin{table}[t]
\centering
\caption{$\mu$ value estimation method comparison}
\begin{tabular}{c|c|c}
 \hline
 &  Methods &  $\mu$ estimation method \\ [5pt]
 \hline 
 
 1& UTA & $\mu_t(s)=\frac{1}{N}\sum_{i \in N^2_w}\emph{y}_t(i)$  \\ [5pt]
 \hline
 
 2& NLTF & $ \mu_t(s)=\frac{1}{N_2N^2_1}\sum_{n=1}^{N_2}\sum_{k} \emph{y}_t(s+k)$ \\ [5pt]
 \hline
 
 3& ANLTF & $ \mu^{\text{NL}}_t(s)=\frac{1}
    {\sum_{i \in N^2_w}\omega(s,i)}\sum_{i \in N^2_w}
    \omega(s,i)\emph{y}_t(i+k)$ \\[5pt]
  \hline   
  
 4& PATF&    $\hat{\mu}_t(s)=\frac{1}{\sum_{t'=1}^{M}\omega(y_t,y_{t'})}\sum_{t'=1}^{M}\omega(y_t,y_{t'})\emph{y}_{t'}(s)$\\[5pt]
 \hline
\end{tabular}
\label{tab:noisefreeEst}
\end{table}

\subsection{Experimental data}

\begin{figure}[t]
\graphicspath{{results/}}
   \centering
\begin{tabular}{cc}
\includegraphics[width=4cm]{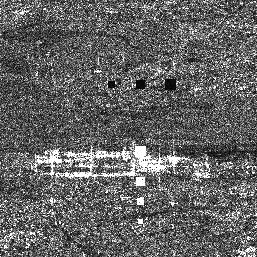}&
\includegraphics[width=4cm]{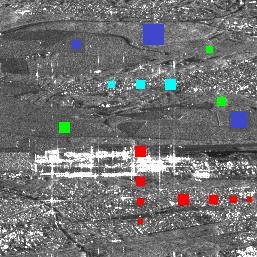}\\
  (a) Noisy image 1 &(b)  Reference map\\
\end{tabular}
	\caption{Realistic simulated SAR image one and reference image. The background image is obtained from an average of 69 Sentinel-1 images.  The change types (the corresponding time series are shown in Fig.\ref{fig:SyntheticCC}) are:  {\color{red}red: step change}, {\color{green}green: impulse change}, {\color{blue}blue: cycle change} and {\color{cyan}cyan: complex change}. This simulated time series will be used for the change classification.}
	\label{fig:SyntheticData1}
\end{figure}

Unlike optical images, SAR images always contain a small number of bright points characterized by local structures \cite{oliver2004understanding}.  To better simulate SAR images, simulated Sentinel-1 data were created based on the arithmetic mean image.
To acquire a round evaluation of the denoising methods, dedicated sequences are created over various areas: forests, farmlands, building areas, etc. Different temporal changes are simulated as shown in  Fig.\ref{fig:SyntheticData1}. Changed values are extracted from the corresponding real SAR time series. All the pixels in the same rectangular at time $t$ have the same noise-free value.

\begin{figure}[t]
\graphicspath{{results/}}
   \centering
\begin{tabular}{c}
\includegraphics[width=7cm]{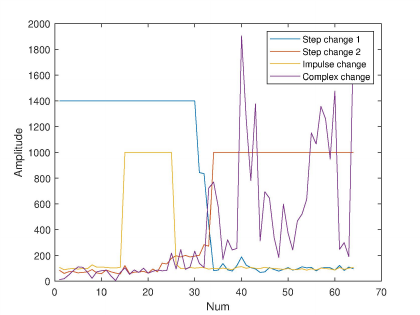}\\
(a) Building area changes \\
\includegraphics[width=7cm]{NoiseFreeNoisy_very_high_res.pdf}\\
(b) Cycle changes\\
\end{tabular}
	\caption{Different time series changes introduced in figure \ref{fig:SyntheticData1}.}
	\label{fig:SyntheticCC}
\end{figure}

64 simulated time series without change and 64 simulated Sentinel-1 images with changes are used to compare different multitemporal denoising methods (UTA, NLTF, ANLTF and PATF). When fewer time series are available, UTA and PATF can use spatial adaptive denoising to improve the temporal denoising results.

Sentinel-1 single look intensity SAR data and Sentinel-1 GRD data are also used for comparison and residual evaluation.



\subsection{Despeckling results comparison with simulated SAR data}

\begin{table*}
\centering
\caption{Residual evaluation results (best value in boldface).  M-index is evaluated in the automatically selected homogeneous area, while residual (Eq. (\ref{scorenormalized})) is calculated based on the whole image. The larger PSNR and MSSIM represent better results, while smaller M-index and residual values represent better results.}
\begin{tabular}{c|c|c|c|c|c}
\hline
 Used data&  Methods &UTA&NLTF&ANLTF&PATF \\ \hline
Denoised data and noise-free image&PSNR&11.790&6.276&7.328&\textbf{24.389}\\\hline
Denoised data and noise-free image&MSSIM& {0.908}&0.769&0.854& \textbf{0.954}\\\hline

Homogeneous area&M-index&3.738&\textbf{1.542}&5.673&2.738\\\hline
Ratio of noisy data and denoised data&Residual&1.025&1.164&1.013&\textbf{0.988}\\\hline
\end{tabular}
\label{tab:numericalEvaluation}
\end{table*}

\begin{figure*}
\graphicspath{{results/}}
   \centering
\begin{tabular}{cccc}
\footnotesize{\rotatebox[origin=l]{90}{(a) UTA with 5$\times$5 window}}&
\includegraphics[width=4cm]{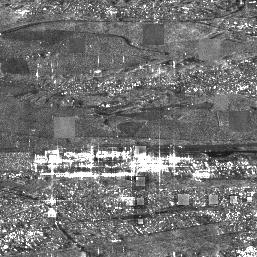}&
\includegraphics[width=4cm]{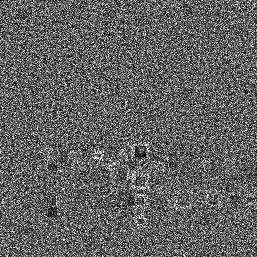}&
\includegraphics[width=4cm]{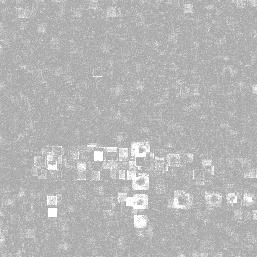}\\
\footnotesize{\rotatebox[origin=l]{90}{(b) NLTF}}&
\includegraphics[width=4cm]{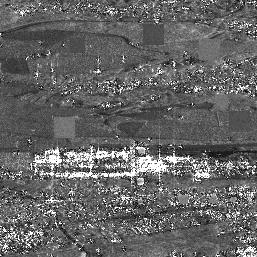}&
\includegraphics[width=4cm]{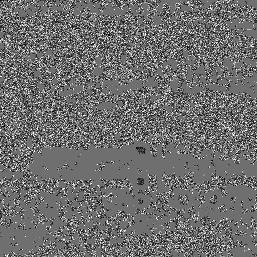}&
\includegraphics[width=4cm]{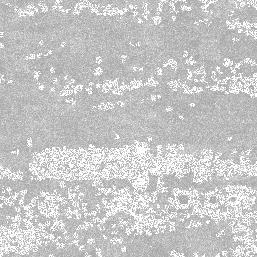}\\
\footnotesize{\rotatebox[origin=l]{90}{(c) ANLTF}}&
\includegraphics[width=4cm]{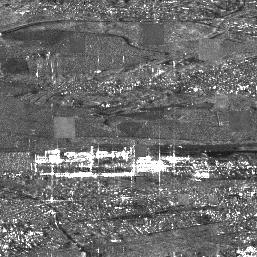}&
\includegraphics[width=4cm]{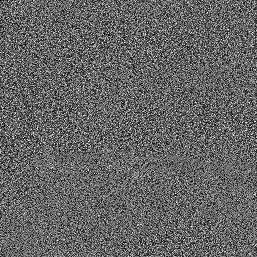}&
\includegraphics[width=4cm]{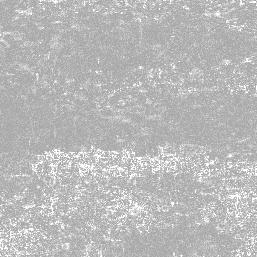}\\
\footnotesize{\rotatebox[origin=l]{90}{(d) PATF}}&
\includegraphics[width=4cm]{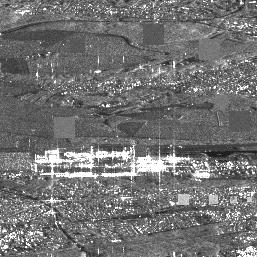}&
\includegraphics[width=4cm]{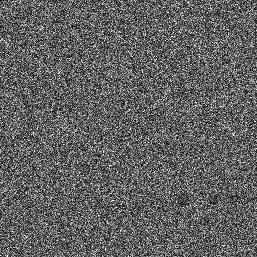}&
\includegraphics[width=4cm]{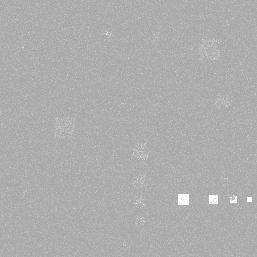}\\
\end{tabular}
\caption{Denoising performances comparison based on 64 simulated Sentinel-1 images. Left: denoising results, middle: ratio with noisy data,  right: ratio with noise-free image. The max ratio values between the denoised data and the noise-free image are used.}
	\label{fig:UTAwithAdaptive4methodsCH}
\end{figure*}

 Using the sliding window to estimate the noise-free value is very easy, but UTA will blur fine features and minor edges (Fig \ref{fig:UTAwithAdaptive4methodsCH}(a)). Compared with the case where adaptive filtering is used, using a rectangular window results in resolution broadening. Because of high-value bright targets in the SAR image, the arbitrary spatial average also causes low-value rectangles in the ratio image. In addition, farmland areas are also not smooth enough.  However, this shortcoming can be avoided by using spatial adaptive denoising (Fig.\ref{fig:UTAwithAdaptive4methodsCH}(c)).

 Visually, PATF provides better results (Fig.\ref{fig:UTAwithAdaptive4methodsCH}(d)). Their denoising results have high spatial resolution and good results for unchanged areas. It also provides the best PSNR and residual evaluation values, as shown in Tab.\ref{tab:numericalEvaluation}. However, it provides biased results in some changed areas because of the used threshold.
NLTF  detects bright points in advance and prohibits any denoising around them.  This step may help keep the original backscattering values of some important objects, but it will lead to noisy building areas. This causes NLTF to provide the smallest PSNR value.
Although it cannot give good results in building areas, it provides the best M-index \cite{gomez2017unassisted} values in homogeneous areas (Tab.\ref{tab:numericalEvaluation}).

\subsection{Despeckling results comparison with SAR data}

\subsubsection{With long time series}

To test the denoising performance of large amounts of data, we take 339 geocoded Sentinel-1 GRD images as an example. 
In this case, using adaptive denoising will be too time-consuming. We only compare  UTA, NLTF and PATF methods.

Based on the previous analysis, the GLR based similarity estimation method without iteration ($h'_{1st}=\infty$) is used to average the data. In this section, the threshold is calculated using a simulated speckle with $\tau= \mathrm{quantile}(d_{1st}(s_t,s_{t'}),\alpha)$.  A similarity
window of size $7\times7$  is used for the single-look SAR images. Spatial adaptive denoising is not used after the patch-based adaptive temporal average.

\begin{figure*}[t]
\graphicspath{{results/}}
   \centering
    \begin{tabular}{cccc}
        \footnotesize{\rotatebox[origin=l]{90}{(a) Noisy and AM images}}&
    \includegraphics[width=4cm]{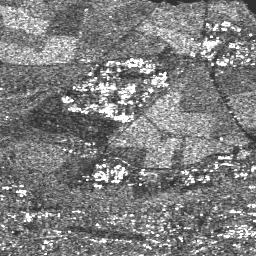}&
    \includegraphics[width=4cm]{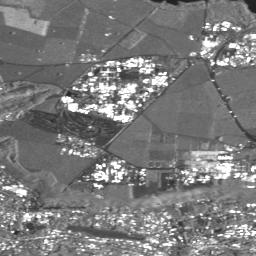}&    \\
    
    \footnotesize{\rotatebox[origin=l]{90}{(b) UTA}}&
    \includegraphics[width=4cm]{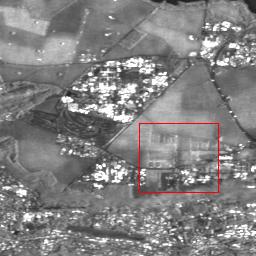}&
    \includegraphics[width=4cm]{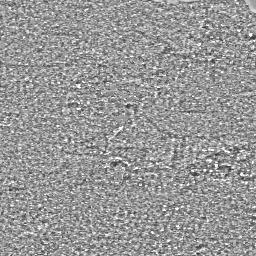}&
    \includegraphics[width=4cm]{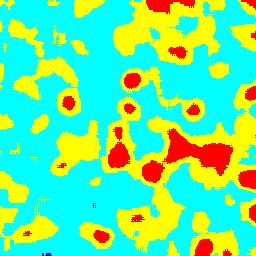}\\
    
    \footnotesize{\rotatebox[origin=l]{90}{(c) NLTF}}&
    \includegraphics[width=4cm]{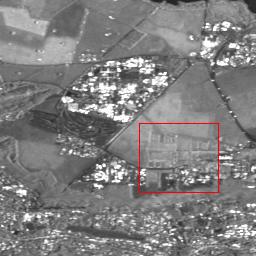}&
    \includegraphics[width=4cm]{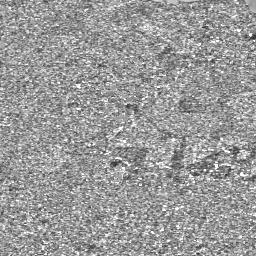}&
    \includegraphics[width=4cm]{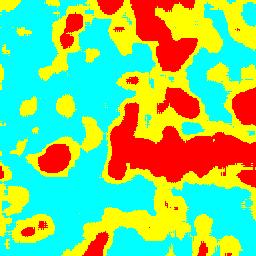}\\
    
    \footnotesize{\rotatebox[origin=l]{90}{(d) PATF}}&
    \includegraphics[width=4cm]{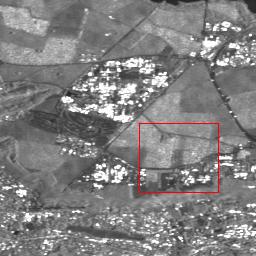}&
    \includegraphics[width=4cm]{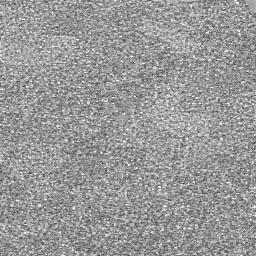}&
    \includegraphics[width=4cm]{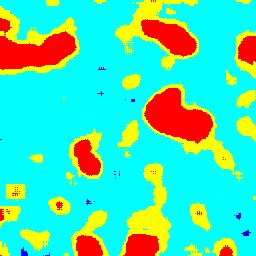}\\
    \end{tabular}

	\caption{Different denoising results comparison based on 339 Sentinel-1 GRD images acquired over CentraleSup\'elec area: denoising results (left), its ratio with noisy data (middle) and residual evaluation results (right).  Buildings are appearing in the red rectangle area. Google Earth Engine is used to prepare the time series data.}
	\label{fig:ComparisonlongTs2}
\end{figure*}

The denoising results are shown in Fig.\ref{fig:ComparisonlongTs2}.  The residual evaluation results have been classified into four classes displaying value ranges [0, 4], and the red colour represents the larger values. All of them can keep the spatial resolution and suppress the speckle noise.
Visually, PATF provides better denoising results than others.  UTA and NLTF provide bias in the bright building area when considering the ratio results. According to the ratio and residual evaluation results (red colour indicates poor denoising) (Fig.\ref{fig:ComparisonlongTs2} (b-c)), UTA and NLTF provide results that have much more wrong denoising results in the changed building areas.  PATF seems to still contain some bias for farmland areas because of their seasonal change (low change magnitude). Empirically, the changes from vegetation to building can be well detected.

\begin{table}
\small
\centering
\caption{Residual evaluation results (best value in boldface).  M-index is evaluated in the automatically selected homogeneous area, while the residual is calculated based on the whole image. }
\begin{tabular}{c|c|c|c|c}
\hline
 Figures&  Methods &UTA&NLTF&PATF \\ \hline
\multirow{2}{*}{Fig.\ref{fig:ComparisonlongTs2}}
&M-index&38.409&4.045&\textbf{2.181}\\
&Residual&\textbf{2.537}&2.970&2.723\\\hline
\end{tabular}
\label{tab:longTSresidual}
\end{table}

\begin{figure*}[t]
\graphicspath{{results/}}
   \centering
\begin{tabular}{cccc}
\footnotesize{\rotatebox[origin=l]{90}{(a) 2SPPB}}&
\includegraphics[width=4cm]{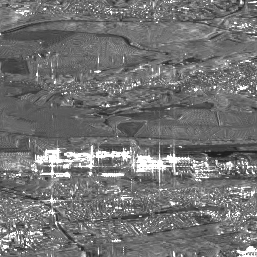}&
\includegraphics[width=4cm]{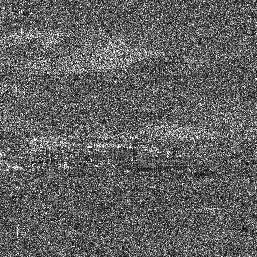}&
\includegraphics[width=4cm]{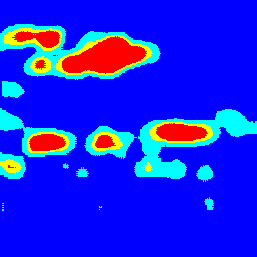}\\

\footnotesize{\rotatebox[origin=l]{90}{(b) MSAR-BM3D}}&
\includegraphics[width=4cm]{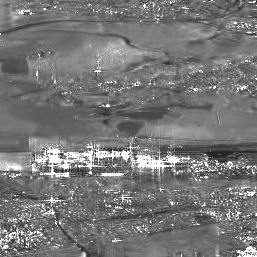}&
\includegraphics[width=4cm]{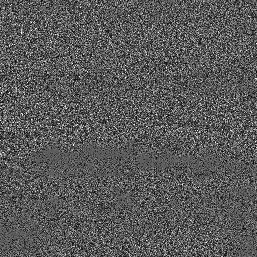}&
\includegraphics[width=4cm]{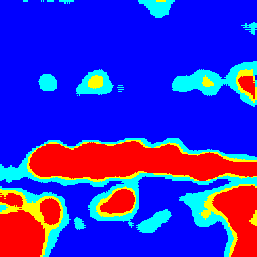}\\

\footnotesize{\rotatebox[origin=l]{90}{(c) RABASAR-DAM}}&
\includegraphics[width=4cm]{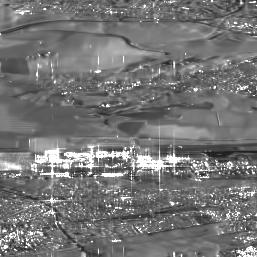}&
\includegraphics[width=4cm]{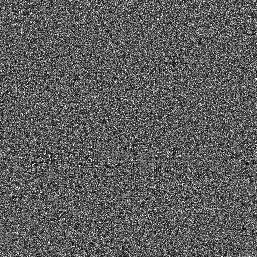}&
\includegraphics[width=4cm]{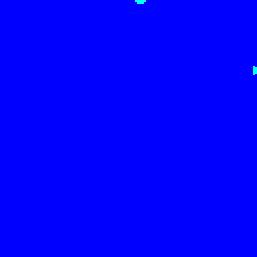}\\

\footnotesize{\rotatebox[origin=l]{90}{(d) PATF}}&
\includegraphics[width=4cm]{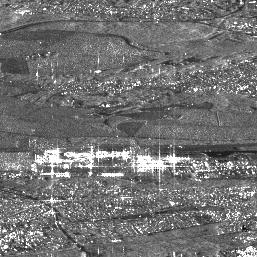}&
\includegraphics[width=4cm]{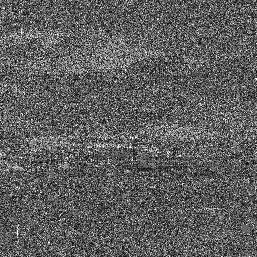}&
\includegraphics[width=4cm]{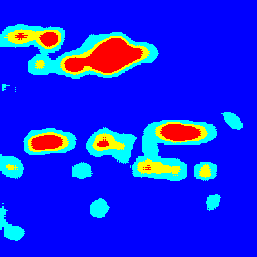}\\

\end{tabular}
	\caption{Different denoising results comparison based on 64 resampled (step 2) Sentinel-1 images: denoising results (left), ratio with noisy data (middle) and residual evaluation results (right).  There are changed building areas in the middle of the image. The residual evaluation results have been classified into four classes with a display value range [0, 4], and the red colour represents the larger values. }
	\label{fig:ComparisonSent}
\end{figure*}


\subsubsection{Comparison with state-of-the-art methods}

Simulated SAR data has i.i.d. speckle, which is not true of real SAR data. Therefore, we expect different results with real data. In this section,  Sentinel-1  SAR images are used for comparison. Spatial and temporal correlations exist in these images. Since there are too many temporal changes for Sentinel-1 images, a 30$\times$30 window is used for the ENL estimation.

RABASAR provides the best results when using Sentinel-1 images.
Figure \ref{fig:ComparisonSent} shows that 2SPPB and PATF provide biased results in the farmland areas (bright areas in the ratio images of Fig.\ref{fig:ComparisonSent}(a) and (d)). This also explains their similar residual evaluation results. When testing the 64 Sentinel-1 data, a weighted average seems slightly better than using binary weight in the bias area.  
PATF can show its advantage with long-time series.

\begin{table}[t]
\small
\centering
\caption{Residual evaluation results (best value in boldface) of Figure 6 }
\begin{tabular}{c|c|c|c|c}
\hline
  Methods &2SPPB&MSAR-BM3D&RABASAR&PATF \\ \hline
M-index&3.501&\textbf{1.931}&13.435&3.371\\
Residual&1.498&2.728&\textbf{0.969}&1.403\\\hline
\end{tabular}
\label{tab:RealSARSentTerr}
\end{table}

    Like NLTF temporal denoising approaches, MSAR-BM3D \cite{chierchia2017multitemporal}  processes the temporal similarity blocks. Temporal filtering is made in the transformed domain. To avoid the $\emph{rare patch effect}$ phenomenon, isolated bright targets are selected and prevented from denoising. Some of the changed points and their adjacent points could also be detected as targets, which may cause noisy pixels and degrade the appearance of denoising results. It is also why the ratio results with noisy data in these areas have low variance.

    Compared with 2SPPB and MSAR-BM3D methods, RABASAR always provides better results concerning the residual evaluation results and the ratio with the noisy image. There is still obvious residue texture in the ratio results of  MSAR-BM3D. When tested with Sentinel-1 SAR data, the denoising results (Fig.\ref{fig:ComparisonSent}) show that MSAR-BM3D provides the best denoising results (also provides the smallest M-index value) in farmland areas without introducing any clear bias.

The residual evaluation results acquired using single-look intensity data seem much better than those using Sentinel-1 GRD data.  Sentinel-1 GRD data is prepared using Google Earth Engine. Sentinel-1 GRD images are preprocessed in that platform by SNAP\footnote{http://step.esa.int/main/toolboxes/snap/} with thermal noise suppression, radiometric calibration, and terrain correction. To save memory space,  the platform converts the float32 values to unsigned 2-byte uint16 integers and retains only the values between the 1st and the 99th average percentile of the values before applying the quantization. This will modify the original SAR signal statistics \cite{koeniguervisualisation}.

\section{Conclusion}

In this paper, we proposed a patch-based adaptive temporal filter and a denoising performance evaluation method.  It acts like a general scheme that can introduce temporal information into spatially based SAR image denoising methods.
Different similarity methods could be used to acquire the temporal similar points. When there is no change and the number of images is large enough, the arithmetic mean of the intensity time series could provide better denoising results. However, when some changes exist, the arithmetic mean introduces artefacts in the changed areas. A patch-based adaptive temporal average could provide better denoising results, especially for the long-time series data. Considering the real temporal SAR images, changes are unavoidable because of dielectric and geometrical property changes in the scattering elements. With the noncontinuous weight calculation function,  PATF can not avoid the bias in the low change magnitude areas.
Thus, temporal change detection despeckling methods could guarantee a good speckle reduction capability regarding PSNR, MSSIM and perceived image quality. Although the denoising performance of PATF does not overcome RABASAR with a limited time series, the simple algorithm allows us to apply it to the Google Earth Engine platform easily.
The residual textures in the denoising results can be highlighted with the residual evaluation method.

\section*{Acknowledgment}

\ifCLASSOPTIONcaptionsoff
  \newpage
\fi

\small
\bibliographystyle{IEEEtran}
\bibliography{IEEEabrv,IEEEexample}

\end{document}